\documentclass[twocolumn,twoside,10pt]{article}
\usepackage{graphicx,amsmath,amssymb,amsthm}
\usepackage[latin1]{inputenc} 

\makeatletter \oddsidemargin-.25in \evensidemargin-.25in
\makeatother \topmargin-0.75in \textwidth7in \textheight9.5in

\theoremstyle{definition}

\begin{document}

\date{}

\title{\begin{flushleft}
  \noindent {\small {\it Journal of Nonlinear Systems and Applications
  (2010)
  \\
  Copyright $\copyright$ 2010 Watam Press \hfill http://www.watam.org/JNSA/\\[5.0mm]}}
  \end{flushleft}
  \Large\bf \uppercase{A multiagent urban traffic simulation} }
  \author{
    Pierrick Tranouez, Éric Daudé and Patrice Langlois
    \thanks{The authors would like to thank the GRR SER and the region Haute-Normandie for the funding of the MOSAIIC program from which this work stems from.}
    \thanks{Pierrick Tranouez is with Litis, Rouen University. (e-mail: Pierrick.Tranouez@univ-rouen.fr)}
    \thanks{Patrice Langlois and Éric Daudé are with UMR IDEES, Rouen University (email: Patrice.Langlois@univ-rouen.fr and Eric.Daude@univ-rouen.fr)}
  }
 \maketitle


{\footnotesize \noindent {\bf Abstract.}  We built a multiagent simulation of urban traffic to model both ordinary traffic and emergency or crisis mode traffic.\\
This simulation first builds a modeled road network based on detailed geographical information. On this network, the simulation creates two populations of agents: the Transporters and the Mobiles. Transporters embody the roads themselves; they are utilitarian and meant to handle the low level realism of the simulation. Mobile agents embody the vehicles that circulate on the network. They have one or several destinations they try to reach using initially their beliefs of the structure of the network (length of the edges, speed limits, number of lanes etc.). Nonetheless, when confronted to a dynamic, emergent prone environment (other vehicles, unexpectedly closed ways or lanes, traffic jams etc.), the rather reactive agent will activate more cognitive modules to adapt its beliefs, desires and intentions. It may change its destination(s), change the tactics used to reach the destination (favoring less used roads, following other agents, using general headings), etc.
We describe our current validation of our model and the next planned improvements, both in validation and in functionalities.}

\vskip.2in


\section{Introduction}
One of the major interrogations in a catastrophe is: will people react correctly to preserve their own security, and doing so, preserve the security of all? This question is most of the time linked with the idea of \emph{risk culture} and \emph{risk training}, which are supposed to produce safe reactions at the right time. Is risk education therefore a key factor to preserve life? Thinking so, we assume that the description of behaviors in risk situations should be engaged in term of personal dispositions, which are mostly related to education. In society, norms such as the highway code, or safety instructions in risk situations, are supposed to govern citizen's behaviors. Do circumstances or local contexts produce different behaviors than those acquired, and then alter global vulnerability? The relation between inherited-based behaviors, which can be viewed here as a "follow the rule" behavior, and the circumstances-based behavior, is an interesting perspective to study the question of the vulnerability of populations.\\

Space is an important factor of risks situations, not only as a support of activities and populations, but also as an actor in itself of the situation. Risk is space related. One defines risk as a probability of space-time interactions between a source (ex: an industrial plant) and a target (ex: an inhabitant building) \cite{Daude1, Daude2}. Space can produce many different contexts throughout its organization. The main difficulty to characterize and manage the risk is then the huge amount of interactions that links entities. Risk has spatial impacts : consider the real estate value near or far from a chemical factory. Risk management transforms the environment, it can be perceived (e.g.  type of allowed constructions, fences etc.). Risk sometimes outlast their management : contaminated soil can remain contaminated decades after the factory generating the contamination has disappeared, along with its most active risk management. This is all the more important in urban area which concentrate a high number of activities, and then reveals some tensions between them. The sharing of a same resource, space, implies some regulations tools in term of laws, of infrastructures and of behaviors. Risks are multi-layered and imply different kinds of actors, human and non-human. A catastrophe in an urban context implies a large number of people and groups, each endowed with their own skills, behaviors and resources. Efforts of public services made in order to mitigate the outcome of the catastrophe are different from, sometime even in contradiction with, those followed by individuals to save their life, but they occur at the same time and the same place. And the same person can respond differently to an event, depending of the context. Furthermore, risks are dynamic, feedbacks and nonlinearities are important. One can observe a domino effect as an explosion in one site produces secondary accidents in the neighborhood, due to the high concentration of activities. \\

The MOSAIIC  project \cite{Tranouez, Daude2} aims to observe and understand local and global effects of individual behaviors in the dynamic of a transportation network system after an industrial accident. Few researches take into account the behaviors of group or individual when studying the risk at the scale of a city. Physical aspects override the measure of risk and population damage is a result of these major forces. In this way, intensity of a toxic cloud or of earthquake defines buffers that are used to estimate the number of inhabitants and equipment implicated in the event, and then gives an estimation of the vulnerability of populations. Traditionally, the population dynamics, its ability to move during a catastrophic event are not considered to evaluate vulnerability of this population!\\

When human behaviors are investigated in a risk situation, it is mostly at a very fine scale, for example rooms or building \cite{Helbing1}, and one kind of behavior is particularly studied: panic and escaping \cite{Helbing2}. When macro scale is preferred, human behaviors are replaced by traffic behavior. Macroscopic models consider parameters such as traffic density or traffic flow to compute road capacity and then the distribution of traffic in the road network. In risk management, many studies have shown that early stages of the phenomenon are critical on the level of the global damage. This situation appears for example when the diffusion of panic between some individuals produce a snowball effect on the entire population. This social contagion assumes hypnotic effects and selfish behaviors \cite{Brown}. But individual behaviors in risk situations are not limited to panic and escaping behaviors, even not to selfish behavior. If one considered Bhopal (1984) or Toulouse (2001) accidents, the number of victims or the time of resilience of the system have largely increased due to curiosity: in some circumstances, people want to \emph{see} the damage.  In some other circumstances, one can observe altruistic behavior when self-centered interest would have been expected \cite{Clarke}. The MOSAIIC project aims at modelling and simulating all this complexity. We first have to construct a model of urban traffic which is able to capture the urban pulsation, i.e. the main circulation flows of the city during specific time periods. This first step in the modeling process is supposed to be the more parsimonious as possible in term of related social data. Our hypothesis is that we can use the structure of the network and the norms (highway code) to deduce the flow between different points in space and time. In this agent-based model \cite{Woolridge}, agents-drivers are supposed to behave as reactive agents, following the rules related to their plans. The second step is to produce a cognitive architecture for agent-drivers to manage specific context: congestion, rerouting, actions due to a major accident, in fact everything which is susceptible to modify its initial plan. 

\section{Dealing with human behaviors in a traffic model}
If traffic modeling is nearing a century of age, most of these models belong to Operational Research (OR) problems -- finding an optimal solution balancing various constraints. In these models, roads and road users were abstracted and aggregated, so as to become a flow problem that could then be optimized. They can answer interesting questions in urban or public transport planning \cite{May}.\\

Sometimes considering average response to a problem is not enough for the scientific problem at hand. We are interested in a dynamic modeling of urban traffic. In this kind of problem, the actions of a few can have a definite impact on the global traffic. An accident implicating half a dozen vehicles in a strategic crossroads of a town can create a traffic jam wave that can affect thousands of vehicles. This is the kind of complex phenomenon we would like to be able to model and simulate. Classic OR tools are not well suited to the task.\\

Although we are not the first to make this statement \cite{transims}, models that tried to alleviate this too-large-scale limitation, have mainly tried to use cellular automata for the task. They added some level of individual-based components to their modeling, but still failed to encompass all that could be needed. Cellular automata are eulerian methods -- intelligence is in one place, rules describe the behavior of bits of space. Values linked to the cells seem to simulate the entities of the modeled system, the same way alternatively lit crystals in an LCD display can give the illusion an object moves around a screen. This contrasts with lagrangian descriptions, where entities of an environment are distinguished, and their spatial coordinates are but one of their describing characteristics. Unlike what can be easily simulated in a CA, lagrangian entities have a trajectory: even in a discretized space a la CA, they can for example act according to something that happened n time steps and m space steps before or away, or according to a plan. This cannot practically  be done in a CA. Multiagent systems belong to this latter category of modeling. As we try to build a model with a grain fine to the level of geometrically correct individual vehicle behavior, from which at least town-quarters-level flow disturbance can arise, we believe this technique is the right one for the task. We then have to build a urban network structure to support these mobile agents.

\section{From geometry to topology}
\subsection{Geographical databases}
A Geographical Information System (GIS) is a system designed for creating, storing, analyzing and managing spatial data and associated attributes. Although it contains a relational database, it needs to go beyond what is needed for classical alphanumerical databases to manage geometrical information, which is continuous by nature, as opposed to the discreteness of usual databases. Indeed for example the database cannot contain all the points of two segments in order to compute a possible intersection: other storing and managing methods must be used for the geometric data of the system.\\

A geographic database is generally comprised of layers or coverage overlapping on a same spatial domain. Each layer contains homogeneous spatial features such as the limits of a city, the course of a river, the geometry of a road etc. Each feature is described in two different ways. First the geometric and optionally topological information is stored in different binary files in the base. Second the record description is a line in the record table; it contains different attributes and descriptions of the feature (generally text or numbers).
\subsection{ESRI Shapefiles}
The first step of the constitution of our system is the constitution of a basic layer of geographic database. This layer is built from the importation of shapefiles, a GIS file format popularized by ESRI \cite{ESRI}. In order to build a traffic simulation, we will build our model from data relative to the road network and optionally from other localized information such as living or working areas.
A shapefile is mainly constituted of three files: one contains the attribute table (.dbf), another contains the geometric data (.shp) and the third is an index allowing matching entries of the first with those of the second.\\

A shapefile contains only the geometric description of objects through a collection of 2D or 3D coordinates that represents, according to the layer type, a cloud of points, open polygon lines for networks or closed polygon lines to describe the boundary of surfaces. The topological information, which describes in geomatics the relationships between the geometric entities, such as connections of edges with nodes in a graph or the adjacency between zones in a surface partition, is absolutely not present in a shapefile, and must therefore be computed by our application form the raw geometry of the imported data.\\

To build a realistic representation of the traffic network of an important urban agglomeration able to simulate the circulation of tens of thousand of vehicles, we had to conceive a network layer structure both complex and efficient. Furthermore, as the importation of data was coming from existing data provider such as IGN, NAVTEQ or Tele Atlas, we had to deal with the way each of them modeled things in their solutions.

\subsection{Urban network structure specifications}
A road network is modeled according to specifications that are in part common to any network and in part dependent on decisions made according to the data
\subsubsection{General specifications}
A road network shares the properties of any geographic network. It is constituted of two main geometric entities: lines, linear components, comprised of several shape points, and nodes, point components that join or terminate lines.\\

These two entities are joined in an oriented multigraph $G=(S,A,f)$  where $S$ is the set of vertices, associated to the geometric nodes, while the set $A$ of edges is associated to the geometric lines, while function $f\colon A\to S \times S$  associates to each edge one initial and one final vertex.\\

Unlike most other geographic information layer, a road network may not be planar: two lines can intersect in their planar projection without modeling an intersection in the real world. This happens when these lines are at different altitudes such as in bridges, tunnels, or motorway embranchments.\\

Furthermore, geographic graph are \emph{topological} graph, which differ from usual graphs in that they are associated to one geometric representation, called the embedding of the graph. Only vertices of degree 3 or more are considered to be true vertex, those of degree 2 being seen as shape points, useful for the geometric information they bring, but not "true" connectors. The geometric representation of the graph is always present to the mind of the geographer, which may create misunderstandings with other scientists more used to a more abstract representation of graph, with planar graph rather than plane graph. As previously said, it is also sometimes extended to non planar graph: the geometric information in the shapefile represents in that case the projection on a connected compact 2-manifold of a graph embedded in a connected compact 3-manifold \cite{Preparata} (intuitively: a 3D graph is drawn on a surface).\\

The attribute table associated to the network will contain all the traffic related information, such as the number of lanes, speed limits, sense of travel etc. Nonetheless this information may not be associated to elementary lines or nodes. For example major roads may contain different lines and important roundabout may contain different nodes and lines. We therefore defined the notion of super-nodes that relate to several nodes (and the assorted sub-graph) and super-edges that relate to several edges (and the assorted sub-graph). G is therefore a hypergraph in these conditions. Whether these conditions are met or not depend on modeling decisions made by the data provider.
\subsubsection{Geographic data based specifications}
There are different ways of structuring the geographic information in a shapefile to model a network.
For example NAVTEQ chose in its Navstreets product to create a node for each intersecting link, even if the road they model are not connected. Another layer represents the relative elevation of the entities of this first layer. Both must therefore be used to correctly build the road network in our simulation. Another example is the orientation of the edges, as the links are oriented following another convention (called \emph{Reference nodes}) than what could be used in a shapefile, and the edge must therefore be computed following this convention.

\subsection{Building the topology from the geometry}
Building a topology from the geometric information contained for example in a shapefile depends on the kind of spatial organization we want to represent.
\subsubsection{Planar mesh}
In the case of a surface mesh (ex: limits of countries, of urban areas, of town quarters etc.), we aim at rebuilding the boundaries and the junction nodes between them from closed polygonal chains (a.k.a. polylines). The layer we produce is thereafter structured around a planar multigraph of vertices, edges and faces, and with each oriented edge associated to 2 vertices (initial and terminal) and to 2 faces (left and right).\\
The building algorithm uses a quadtree and a tree connecting each point, in which all the points of the shapefile are organized. Each leaf of the quadtree contains a point $P_{i}$ and 4 branches for the 4 quadrants of space (NE, SE, SW, NW) surrounding $P_{i}$. This structure allows for a quick detection of the multiplicity of points. For example, a point with a multiplicity of 3 or more will be associated to a vertex, while a point of a multiplicity of 2 will be a shape point of an edge. Furthermore, the connection tree allows the quick detection of adjacent points along a polyline, and detecting the superposition of two lines forming the boundaries of two zones, or the succession of angular sectors around a vertex common to three polygons or more.
\subsubsection{Networks}
In order to build the structure of a planar network (for example hydrographic or of roads), we do not store faces but the polar order of succession in the edges. Each edge stores the next edge turning left and the prior edge turning right. This structure is known as DCEL, Doubly Connected Edge List \cite{Preparata}. The algorithm to generate this topology uses the dynamic quadtree structure used that was used to build the DCEL.\\
The road network often exists in 3D, although despite the existence of this possibility in the format specification, most shapefiles only contain a 2D geometric representation. The data provider must in that case model the altitude differently, and our algorithm must be adapted to this. For example NAVTEQ's Navstreets \cite{NAVTEQ} uses another layer called z-levels that must be consulted to know whether a point corresponds to a node or not.\\
At the end of this step, we have a topological graph that is structured like the road network, but without its semantics. We will now build from it and from the database part of the shapefile a non-topological graph that models this ontology.\\

\subsection{From static topology to traffic-oriented}
\subsubsection{Traffic oriented graph}
Our traffic model is individual-based: each vehicle will be modeled as an agent. This implies the creation of an adapted environment for them, in terms both suitable to their ontology and adapted to the geographic data we reaped. For that a graph will be built, a transport graph that will contain the necessary structures and values.\\
This first version of our models is only interested in simulating motor vehicle: pedestrians and bicycle are ignored.\\
The database contains the sense of travel and the traffic restrictions for each topological edge. One oriented edge is created for each sense of direction allowed for motor vehicles.
Edges and vertices of the transport graph are called elements. To each element is associated a data container and a vehicle transporter.\\
The data associated to an edge are for example its geometric length, its number of lanes, its speed limits etc.\\
The data associated to a vertex is notably the size of the container of its transporter, depending on the number and the sizes of the edges connected to him.\\
Transporters are non-mobile agents associated to elements. They handle parts of the collective behavior of the vehicles. They will be described in more depth in the following section.
\subsubsection{Routes in the graph}
Mobile agents will try to reach destinations in the graph. As we intend to simulate a realistic traffic of tens of thousand of vehicles, we want to facilitate their computing of their trajectory. To do that, we build a set of "shortest" path stored in the traffic graph.\\
We compute a weight on the edges that combines different parts of its data: its length, the speed it can reasonably be driven upon, its estimated width based on the number of lanes etc. to model the attractiveness of this edge. After that we compute Dijkstra's algorithms \cite{Dijkstra} from each vertex to all the others, which we store in each vertex. This data takes $numberOfVertices^{2}$ bytes of data, which is important, but allows the computation of a good path by an agent in constant time, which is a good thing as hundreds of agents are generated at all time in the simulation (simulating vehicles entering the road network of the simulated urban agglomeration).

\section{Mobile agents on the network}
Our agents are mainly so far car agents, trying to go from one place to another.
\subsection{Strategic behavior}
Modeling in details the various detailed trajectories of car users is a research problem in itself \cite{Banos}. Nonetheless we are not interested in who did what or why, but only in what are the fluxes in our network in typical scenarios. When an agent is injected in the network, a starting point and a destination are randomly chosen. \\
This randomness is not necessarily uniform. If we suppose the agglomeration centered on its main town, like the agglomeration of Rouen that we simulated more than others, we can shape different distribution, favoring the likelihood of drawing rather a inner or an outer edge for example. Traffic between 8:00 AM and 9:00 AM for example starts mainly on the border or outside the agglomeration and ends to the same distance to the center (outer edges): we can simulate traffic that do that. When shops close in the town center, we have a traffic that is mainly outer bound, with a more important center generation: we can simulate that. We do not have to know what this car and its driver did in the morning, we don't have to simulate realistically its history, as long as we model the actual traffic fluxes right.
\begin{figure}[htbp]
\begin{center}
\includegraphics [width=\linewidth]  {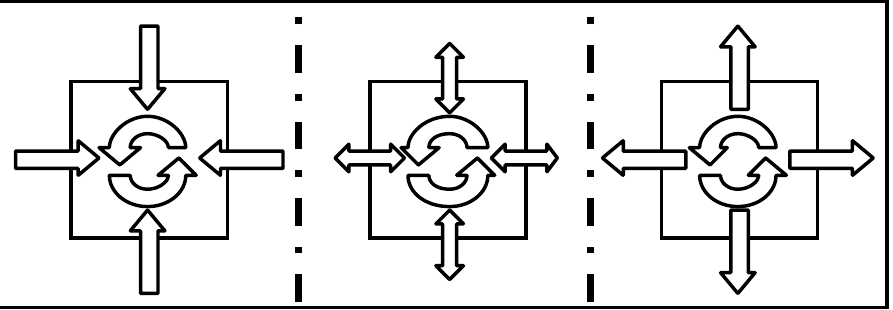}
\caption{3 different scenarios of source/destination pseudo-random choice}
\label{source-destination}
\end{center}
\end{figure}
Once the agent knows where it is, and what its destination is, it can use the best paths stored in the traffic network to plan a trajectory. It then drives here, adapting his path through its tactical behavior, and managing its immediate surroundings through its operational behavior.
\subsection{Operational behavior}
The planned trajectory of an agent is a succession of edges. Once in an edge the agent tries to drive to its end, the next connection, where it will be able to choose the next planned edge.\\
When it enters an edge, the agent first chooses a lane if several are available, based on the traffic density in each, with a bias for the rightmost lane. As we have a good geometric description of the lane, the driving behavior is fairly detailed, incorporating the length of the car, its capacity/willingness to accelerate and brake, the taste of its driver for long/short safety distance, its taste for following or breaking speed limits etc. All this is incorporated in a driving model inspired by Martin Treiber's Intelligent Driver Model \cite{Kesting}. IDM is a longitudinal traffic model, so we had to expand it to handle multiple lanes and crossroads (the original IDM works for an unlimited one-way, one-lane road) ; we did not use Treiber's MOBIL lane changing model as it is better adapted to motorways than to urban lane changing decisions.\\
The data provided by geographic providers does not include right of passage or traffic lights at crossroads. We therefore had to develop our own model aiming at the simulation of crossroads in a heavy traffic.\\
When a vehicle reaches a crossroad, it slows down and acts according to the fluidity of traffic in the crossroad, in the edge it is currently upon and in the edge it whishes to go to. If they are encumbered, it will more often wait in its way, but it may enter the crossroad and wait here, thus encumbering it (with a more or less strong individual tendency to do so). If the edge it is aiming at has multiple lanes, it will watch both of them, to see if it could fit in one.
\subsection{Tactical behavior}\label{TacticalB}
Although vehicles have an original plan, they will adapt it to what they perceive of their environment. When stuck in what they perceive is a jam, they will try to find alternate routes out of it to their destination.\\
The first method we used is the simpler one. When a vehicle doesn't move enough to its liking -- this saturation is variable amongst agents -- it tries to take alternate paths as soon as possible, favoring the roads with least dense circulation -- although this is not absolute, so as to avoid loops. Once it estimates it's far enough from the jam that sprang this alternate behavior, it resumes its standard behavior, using the best path table to find a suitable one to its destination.\\
The second one is more sophisticated, as it will have uses beyond mere traffic avoidance. Its intelligence is modeled more in the Transporter agents than in the vehicles. Transporters estimate their encumbrance. To do that, they employ direct measure -- how many vehicles do they contain over how many vehicles can they contain in average -- but also statistics on the proportion of vehicles they contain that are annoyed by the traffic -- as described in the first method -- and information from the Transporters around them. If based on this they decide they are encumbered they also warn the Transporters around them of their perception. This will lower the threshold for them to feel encumbered.\\
Once encumbered, the nodes they are connected to will recompute their best path table, using a huge weight for the encumbered edges. When a vehicle arrives to one of these nodes and wants to go to one of the jammed edges, it is informed of the edge state, and it can recompute a route around it, or take the edge anyway.\\
This mechanism is also \emph{theoretically interesting}, as it is an implementation of an emergent property: the interactions of individual behavior affect the behavior of an agent of an higher scale, who alters his behavior, which in turns transforms the behavior of the lower level vehicles. This reifies the perception an individual driver can have of the state and dynamics of the traffic he is plunged in as a whole.\\
A Transporter can also be barred, because of an accident for example. In that case the same mechanism is used, except that this time circumnavigating is mandatory.
The mechanism of these two states is especially useful in what was the original purpose of our model and its main application: simulating urban important accidents -- such as industrial accident -- as the modeler can bar the edges it wants as part of his scenario, and see how the traffic adapts to it in simulation real time, as the vehicles discover the evolving road network and fluxes.

\section{Unexpected individual behaviors in risk situations}
MOSAIIC is about situations where contextual mobility can occur and can diffuse or have large consequences on the global circulation. We call contextual mobility a mobility which is associated to short-range goals (for example to avoid a crowd or a jam) and whose results differ from the initial planning (for example changing the planned trajectory). We will now consider an urban industrial accident. This accident has a finite extension area and a well-determined intensity, represented by a buffer. Inside this buffer, a proportion of people, related to intensity for example, want to change their initial plan. Outside this buffer, behaviors are less reactive. Some want to escape, others want to see and for others "show must go on", and they want to follow their way. In this example we see that the same person might have different behaviors depending on its location related to the event, or related to the behaviors of its neighbors. 
\subsection{Contextual mobility}
We have then defined different kinds of behaviors and methods related to different short-term goals:
\begin{itemize}
\item Chicken behavior: the goal is to find the opposite direction of the source (the buffer), and to drive following this way;
\item Spectator behavior: the goal is to find the source of danger and to go there. If agent is already in the place, then he stays here;  
\item Pragmatic behavior: here the agent selects a new destination in the network and tries to reach it. This behavior simulates the fact that some people will want to reach their children at school or husband or wife at their working place;
\item Wandering behavior: there is no goal, this behavior is the sign of distress. At each time step, just select a road and go there.
\item Roadrunner behavior: this method consists in always selecting the less congested road and to go there. This method can be connected to the Chicken or Bystander behavior;
\item Sheep behavior: here agent follows the crowd whatever the direction.
\end{itemize}
We will discuss now how the behaviors themselves can be implemented, not why or when one or the other will be chosen.
\subsection{Behaviors classification}
In order to implement them, we will distinguish three categories of behavior: global, planar and local. These categories are based on the actual behavior, and not on the motivations behind it.\\
A global behavior is one that makes a reasoning about the road network. Pragmatic behavior falls in this category: the agent will try to find a good path to his newly decided destination using his knowledge of the network. Spectator can also fall here.\\
A planar decision also chooses a destination but tries to reach it using orientation as if no roads existed, as if the vehicle was on an open plan. Of course the network will offer constraints, but a general cardinal like direction will guide the agent. Chicken and possibly Spectator will fall in this category. This means there are two possible implementations of this behavior.\\
A local decision is one based on local-only data: Wandering, Roadrunner and Sheep will fall there.\\
\subsection{Class implementation in the simulation}
Global behaviors are implemented in the agents to allow them to reach their initial destination. The motivations change between those two cases, but the underlying mechanics stay the same.\\
Local require little complexity. Wandering is trivial, Roadrunner and Sheep differ only by the sign of their optimization. We also implemented a simple anti-loop measure: Roadrunners for example will choose the less congested road unless they already went recently through this one.\\
Planar require the ability to choose an edge out of a node based on a global direction. Depending on what the modeler desires, he can choose a distance from the current road intersection, and the agent will choose the intersection at less than the selected distance (expressed in Euclidean distance or number of edges in a path leading to it) that is the closest to the desired direction. An anti-loop measure can be added.\\
The behaviors previously described are ways of coping with an extraordinary situation. Most urban important accidents will have their consequences felt locally at the beginning, before it spreads. The evolution of the perturbation will be like waves spreading from the initial locus outward. If the extraordinary behaviors are the waves, the metaphorical medium of this propagation is the ordinary traffic flow. We therefore need a sophisticated modeling of the day-to-day activities of vehicles in an urban agglomeration as we have seen before.
\subsection{Example}
\begin{figure}[htbp]
\begin{center}
\includegraphics[width=\linewidth] {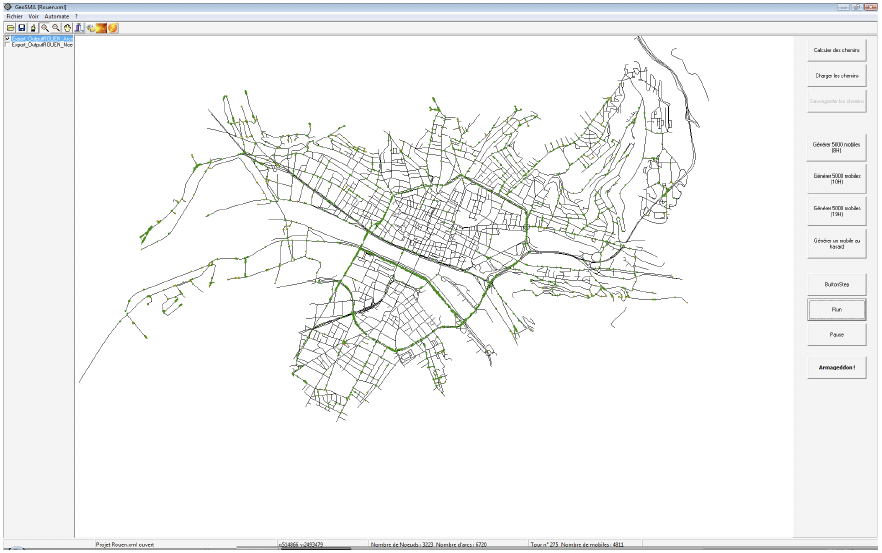}
\caption{An example of traffic in a town before an accident occurs.}
\label{traffic-ordinary}
\end{center}
\end{figure}

\begin{figure}[htbp]
\begin{center}
\includegraphics [width=\linewidth] {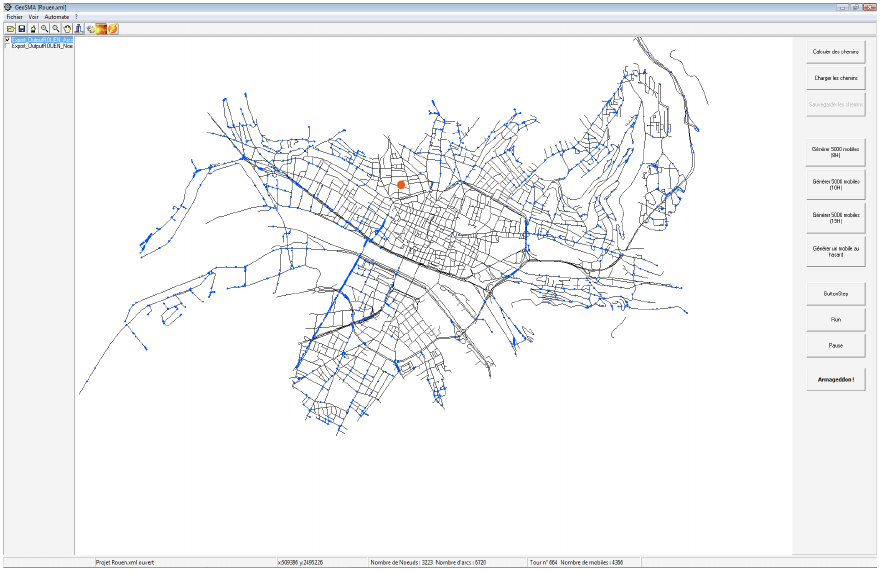}
\caption{The same traffic after the accident occurs (red circle). The agents are all adopting Chicken behavior (in blue).}
\label{traffic-accident}
\end{center}
\end{figure}

Figure \ref{traffic-ordinary} depicts an example of the distribution of vehicles in the main roads of the city. Starting at this point, we generate an event in the city that is supposed to represent an accident. This event, for the purpose of the simulation, is perceived by all individuals and is considered as a repulsive event. In figure \ref{traffic-accident}, this event is a mouse-click event located by the user without any consideration about the reality of the area. As agents perceived the impact zone (in fact its euclidean coordinates), they all change their planned trajectory. Once in a crossroad, all mobiles pick out the Chicken behaviors and pick an edge, choosing the one whose other node forms the widest angle (as close to $\pi$ as possible) with their current node and the coordinates of the explosion.  
This escaping behavior is for instance not applied in concurrence with any other mobility strategies or tactical behavior: they have not the possibility to avoid traffic jam or loops. The main effect of the general application of this rule is purely the draining of the transportation network. Of course this "Hollywood panic" scenario is not relevant in reality but let us test implemented mechanisms.   
Vulnerability increases when a certain quantity of actors changed their dynamics of mobility, mainly after a shift in their goals. Beliefs, desires and goals are then important to take into account in this kind of model.

\section{Discussion}
\begin{figure}[htbp]
\begin{center}
\includegraphics[width=\linewidth] {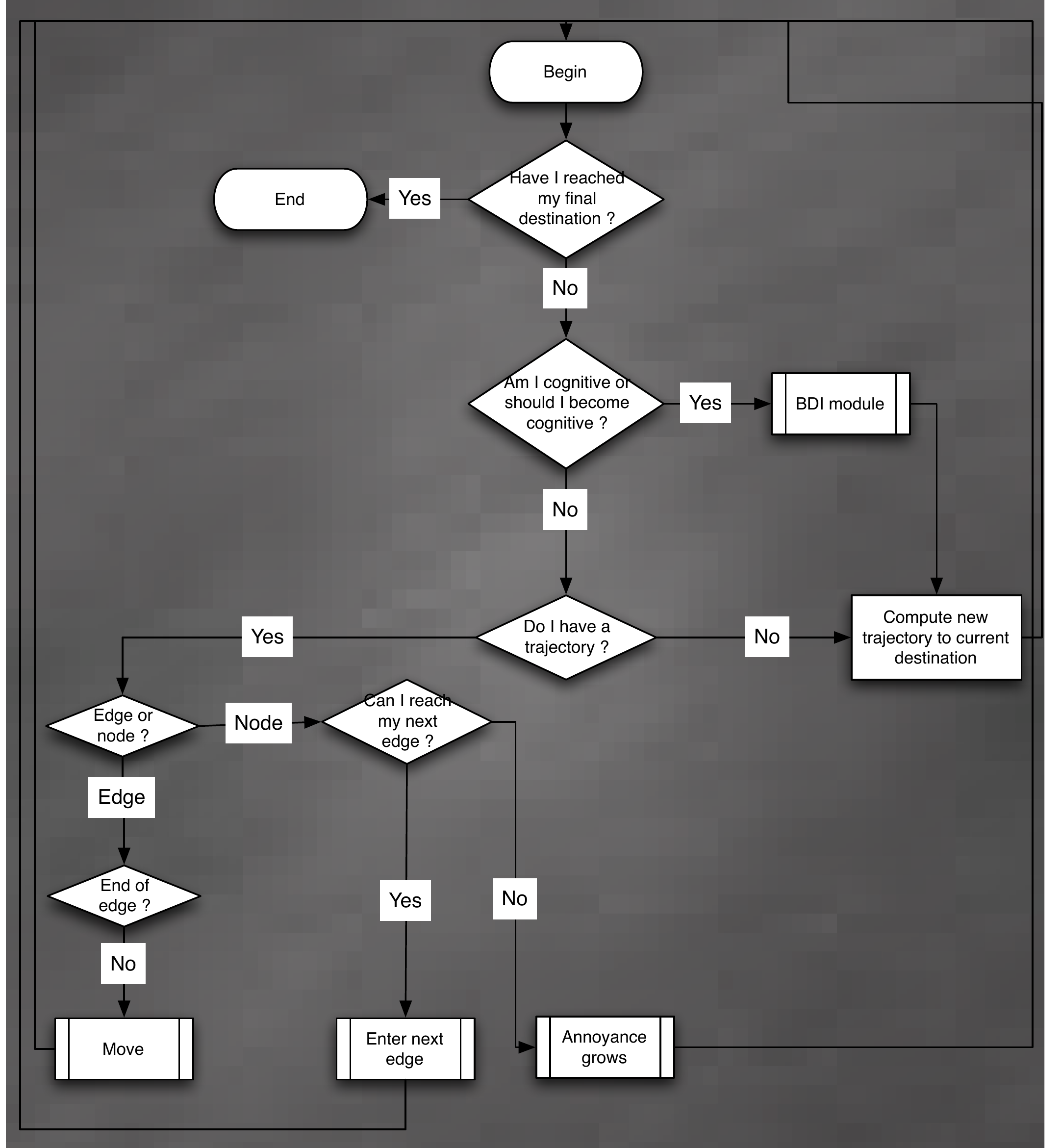}
\caption{How our mobile agents think.}
\label{mobile agents}
\end{center}
\end{figure}

One of the problems for the validation is that modeling as seldom been taken to such a detail level. This level is necessary because of the multi-level nature of traffic: the decision of one driver can start a jam or jams for thousands of drivers, half a town away, half an hour later, a la butterfly effect. Macro model of fluxes, which dominate the field of traffic simulation, cannot do that. Their validation for example is often based on the fundamental diagram of traffic flow of a few selected axes, for hourly traffic. We can compute second by second fundamental diagrams of each edge of our network. We can therefore be an order or two of magnitude more precise in our measure, but what to do of all this information? Indeed, if we describe in details the behavior of vehicles, one must not lose sight that they are not what we are trying to model, the traffic is what we are trying to model, that is their behavior as a group. We fine-tune individual behavior to have the emerging group behavior right.\\
What we have ascertained so far is that:
\begin{itemize}
\item Most edges comply with the Fundamental Diagram made over 5 minutes of time, most of the time. This remains the case even once the measure-time unused edges are taken out of the count We simulated our university home agglomeration of Rouen with up to 50 000 vehicles, and traffic specialists find the results subjectively very satisfying
\item We tried to compare the results of our simulations with data we had about the traffic of the Rouen agglomeration. The data dated from 2001, while the geographic data we had for the network dated from 2006-2007. The western part of the road network had changed too much during this period for any solid conclusions to be drawn from it, despite superficial resemblances in other parts of the networks.
\end{itemize}
A PhD student has recently started a study to increase our knowledge of the traffic in Rouen agglomeration, and thus give us more data to compare our simulation to.

\section{Further developments}
We have defined methods modeling mobility itself, but we now need to model the decision processes for picking or switching between these methods. In an ordinary situation, people follow their own planning and most of the times never deviate of their schedule. But how to justify and explain the fact that in some circumstances, people shift from one behavior to another, from an ordinary behavior to one of the extraordinary described here such as Sheep or Roadrunner?\\
We have so far described models that have been implemented, simulated and whose results are being investigated. We will now describe what we are currently building and improving : providing some cognitive abilities to our agents to model their alternative strategical and tactical decision making.\\
\begin{figure}[htbp]
\begin{center}
\includegraphics[width=\linewidth] {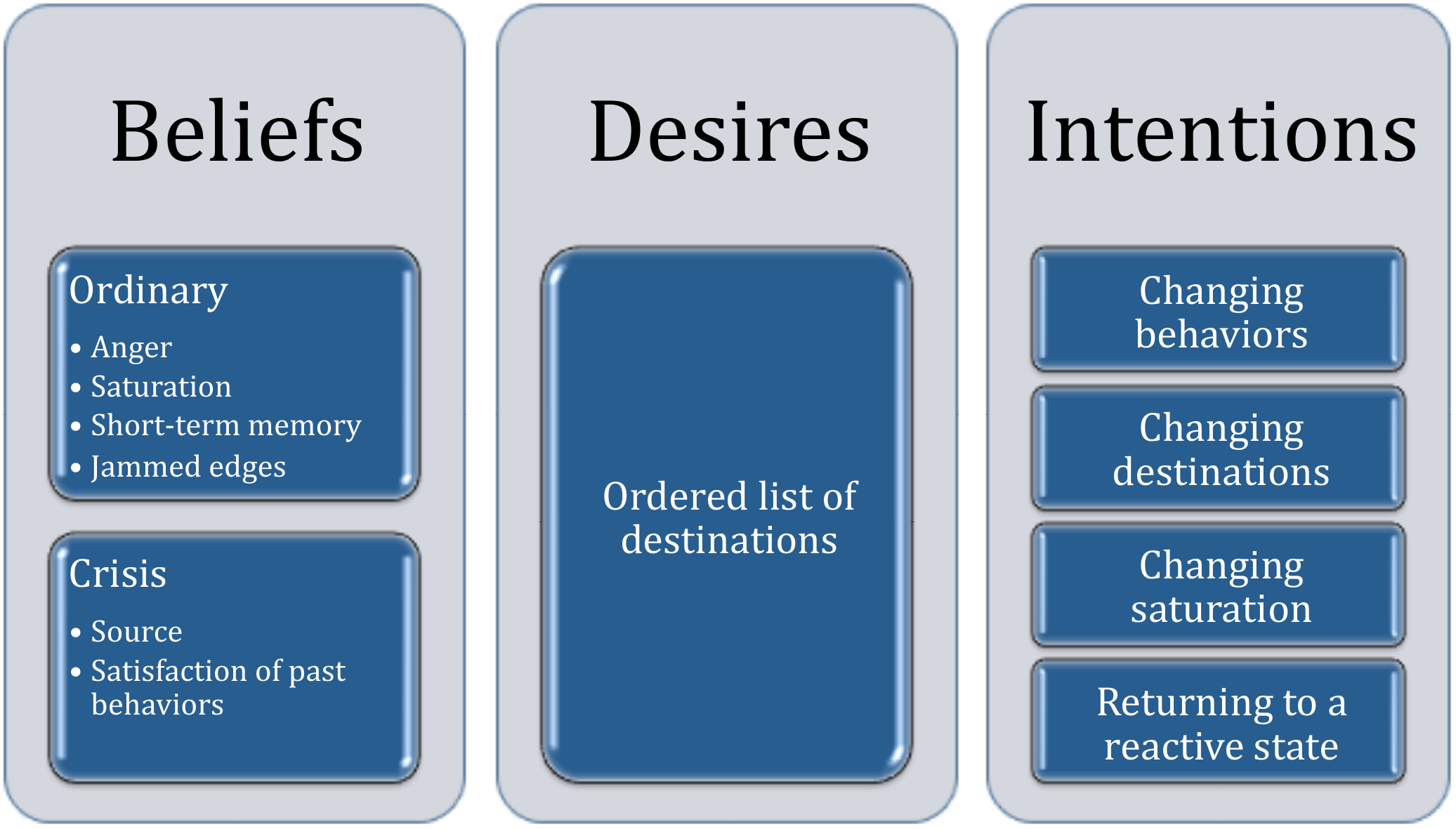}
\caption{Beliefs, Desires and Intentions of our agents.}
\label{BDI}
\end{center}
\end{figure}

We use here a Beliefs - Desires - Intentions (BDI) \cite{Woolridge2} description of the decision process of our agents. Our BDI implementation is simple and homemade : although we need to model some complexity of the decision making, what we need to get right is the behaviour of a large crowd of agents, not necessarily be psychologically realistic at the agent level.
\begin{itemize}
\item Beliefs here model inner and outer perception (am I happy or annoyed, is the circulation around me fluid or jammed), and also acts as a short term memory : 
\subitem Crisis source : If I am in a crisis mode, what caused it, where was it and when did I learn about it ?
\subitem Anti-loop : have I used this road recently ?
\subitem Behavior satisfaction : Has my annoyance grown or shrunk since I started using this behavior ?
\item Desires are a list of destinations. Our mobile agents always want to go somewhere. Initially they have a list of destination (dropping the children at school, fetching the laundry, going to work). This list can be altered in case of a crisis ; ex : I stop trying to go to work, go to my spouse job place to get her then head to my children school, and finally flee.
\item Intentions mode the plans of action of our agents. They know where they want to go, they have a perception of their mental states and their surroundings, now how do they reach their goals ? The tools we give them for that is a possibility of switching their tactical behaviors (cf. \ref{TacticalB}), changing their destinations, and change their saturation (should I often try new tactic or stick to one for a longer time?). 
\end{itemize}

One of the main questions is then: how fine a knowledge and understanding  of people mental procedures is it necessary to have to simulate crowd dynamics and vulnerability of transportation network? In other words, what level of detail is needed in the modeling of individual agent to accurately model the behavior of a crowd of them?

\section*{Acknowledgements}
The authors would like to thank the GRR SER and the region Haute-Normandie for the funding of the MOSAIIC program from which this work stems from.

\footnotesize


\end{document}